\documentclass[10pt,twocolumn,letterpaper]{article}

\usepackage{cvpr}
\usepackage{times}
\usepackage{epsfig}
\usepackage{graphicx}
\usepackage{amsmath}
\usepackage{amssymb}

\usepackage{cite}
\usepackage{amsfonts}
\usepackage{algorithmic}
\usepackage{textcomp}
\usepackage[caption=false,font=footnotesize]{subfig}
\usepackage{booktabs}
\usepackage{tabularx}
\usepackage{multirow}

\usepackage[breaklinks=true,bookmarks=false]{hyperref}

\cvprfinalcopy 

\def\cvprPaperID{****} 

\ifcvprfinal\fi
\begin{document}

\title{Universal Pooling - A New Pooling Method for Convolutional Neural Networks}

\author{Junhyuk Hyun \quad\quad Hongje Seong \quad\quad Euntai Kim\\
 Yonsei University, Seoul, Korea\\
{\tt\small \{jhhyun, hjseong, etkim\}@yonsei.ac.kr}
}

\maketitle
\thispagestyle{empty}

\begin{abstract}
Pooling is one of the main elements in convolutional neural networks. The pooling reduces the size of the feature map, enabling training and testing with a limited amount of computation. This paper proposes a new pooling method named universal pooling. Unlike the existing pooling methods such as average pooling, max pooling, and stride pooling with fixed pooling function, universal pooling generates any pooling function, depending on a given problem and dataset. Universal pooling was inspired by attention methods and can be considered as a channel-wise form of local spatial attention. Universal pooling is trained jointly with the main network and it is shown that it includes the existing pooling methods. Finally, when applied to two benchmark problems, the proposed method outperformed the existing pooling methods and performed with the expected diversity, adapting to the given problem. 
\end{abstract}

\section{Introduction}
\label{1}
Since gaining popularity in the ImageNet Large Scale Visual Recognition Competition (ILSVRC) \cite{b1} of 2012, convolutional neural networks (CNNs) have received much attention from various research fields. A CNN consists of several elements, including convolutional layers, pooling layers, and a fully connected layer. Since CNN first became famous in ILSVRC 2012, pooling has been included in all popular networks: AlexNet \cite{b2}, GoogleNet \cite{b3}, Visual Geometry Group (VGG) \cite{b4}, Residual Neural Network (ResNet) \cite{b5}, and DenseNet \cite{b6}.

Pooling plays several important roles in CNNs. First, it reduces the size of the feature map, enabling training and testing with a limited amount of computation. It also reduces the computation of the network. Second, pooling increases the size of receptive field size in each neuron, enabling each neuron to "see" a larger portion of the input image. This functionality improves the recognition performance. Third, pooling reduces the negative effects of noise and slight distortion. Pooling admits only the representative values and removes the small noises. In essence, pooling simultaneously improves the recognition performance and reduces the computational demands.

The deep learning performance depends on the choice of the pooling method. The most popular pooling methods are average pooling, max pooling, and stride pooling. Overall, these simple methods are efficient because they easily compute the representative values. However, they do not consider the various input patterns and there is still a room to exploit for improving the pooling. Recently, several new pooling approaches have been reported, including a linear combination of simple existing poolings \cite{b15}, and an adaptive pooling method that magnifies the spatial changes and preserves the important structural details \cite{b16}. Most of the new pooling methods assume fixed weights for feature training, which do not account for the characteristics of individual images. Therefore, the fixed-weight constraint on the pooling might degrade the performance of the CNN.

 On the other hands, researches on attention module \cite{b7, b8} have been actively conducted to improve the performance of the network. The attention concept weights a major feature in the feature map. Several studies have shown performance improvements after weighting appropriate parts of the feature maps. Attenuation modules  have been investigated in the spatial domain, time domain, and channel direction.
 
This paper proposes a new pooling method named universal pooling. The proposed pooling is the most general pooling and includes the previous simple poolings such as max pooling, average pooling and stride pooling as special cases. The basic idea of universal pooling is to interpret pooling as attention and extend it to the general channel-wise local spatial attention. That is, the universal pooling selects pooling weights for each channel and they are trained together with other feature extraction parts. Because it embraces the existing simple poolings, universal pooling is expected to outperform the independent simple poolings.
The remainder of this paper is organized as follows. Section \ref{2} reviews some related works, and Section \ref{3} introduces the preliminary fundamentals of pooling. Section \ref{4} proposes the universal pooling concept and discusses its relationship with the simple pooling approaches. Section \ref{5} applies the proposed pooling to two benchmark datasets and presents the experimental results. Conclusions are drawn in Section \ref{6}.

\section{Related Work}
\label{2}
The commonest pooling methods are max and average pooling, but stride pooling in which the feature map is reduced using the stride in the convolution layer has also gained traction in recent years.

Max pooling divides the feature map into blocks and collects the maximum feature value in each block into a smaller output matrix. Max pooling is commonly used between convolution layers and is employed in AlexNet \cite{b2} and VGG \cite{b4}. Average pooling operates similarly to max pooling, but outputs the average of each block in the feature map. Global average pooling (GAP), which applies average pooling over the entire feature map, is commonly used in convolutional networks such as ResNet \cite{b5} and DenseNet \cite{b6}. In DenseNet, average pooling is applied between the convolution layers. Meanwhile, stride pooling is equivalent to importing values from a fixed position after convoluting across the entire area. This approach is adopted in ResNet.

All of these pooling methods are efficient but simple, and it seems that there is a room to improve the performance. S3pool \cite{b9} and stochastic pooling \cite{b19} adopt a probability-based pooling approach. $L_p$  pooling of various coefficients' norms was proposed in \cite{b10} and \cite{b11}, and a fractional version of max pooling was proposed in \cite{b12}. The spectral space was down-sampled through a filter in \cite{b13} and \cite{b14}. In \cite{b15}, the existing simple pooling methods were combined to improve the pooling performance. Detail-preserving pooling, which preserves the feature details by applying existing down-sampling techniques in the image processing area and by learning the parameters of the function over the network, was proposed in \cite{b16}.

\section{Preliminary Fundamentals}
\label{3}
\begin{figure}[htbp]
\centerline{\includegraphics[width=0.9\linewidth]{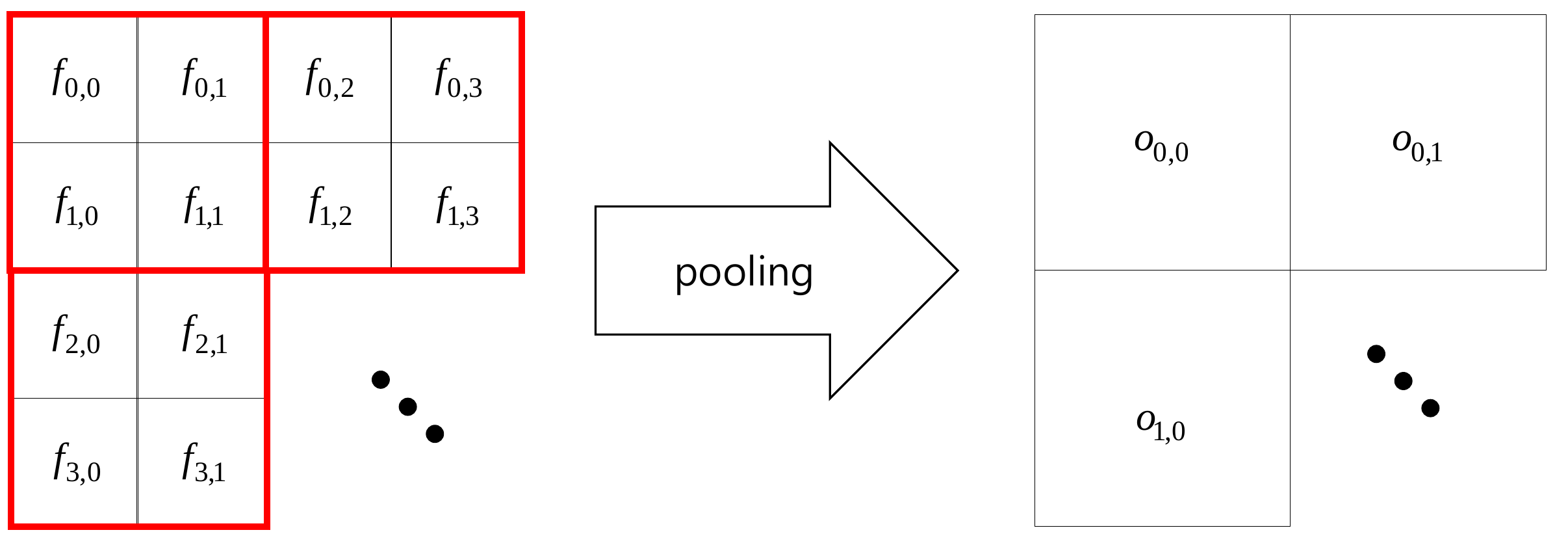}}
\caption{Pooling downsizes the feature map. Each red block denotes one pooling block.}
\label{fig1}
\end{figure}

The pooling operation, which down-samples the feature map, can be considered as a multiple input function
\begin{equation}
\rho :{\Re ^{{S^2}}} \to \Re 
\label{eq1}
\end{equation}
which returns only a scalar value. $S^2$ is the area of the pooling block $\left( S \times S \right)$. For example, let us consider a pooling of size 2 with a stride 2, as shown in Figure \ref{fig1}.


In this case, the pooling can be represented by a multiple input function 
\begin{equation}
{o_{0,0}} = \rho ({f_{0,0}},{f_{0,1}},{f_{1,0}},{f_{1,1}}) .
\label{eq2}
\end{equation}
As mentioned above, there are three popular pooling methods: max pooling, average pooling, and stride pooling. In the following, these methods will be explained on a pooling block of size $S=2$ and stride 2.

\subsection{Max pooling}
\label{3A}
\begin{figure}[htbp]
\centerline{\includegraphics[width=0.9\linewidth]{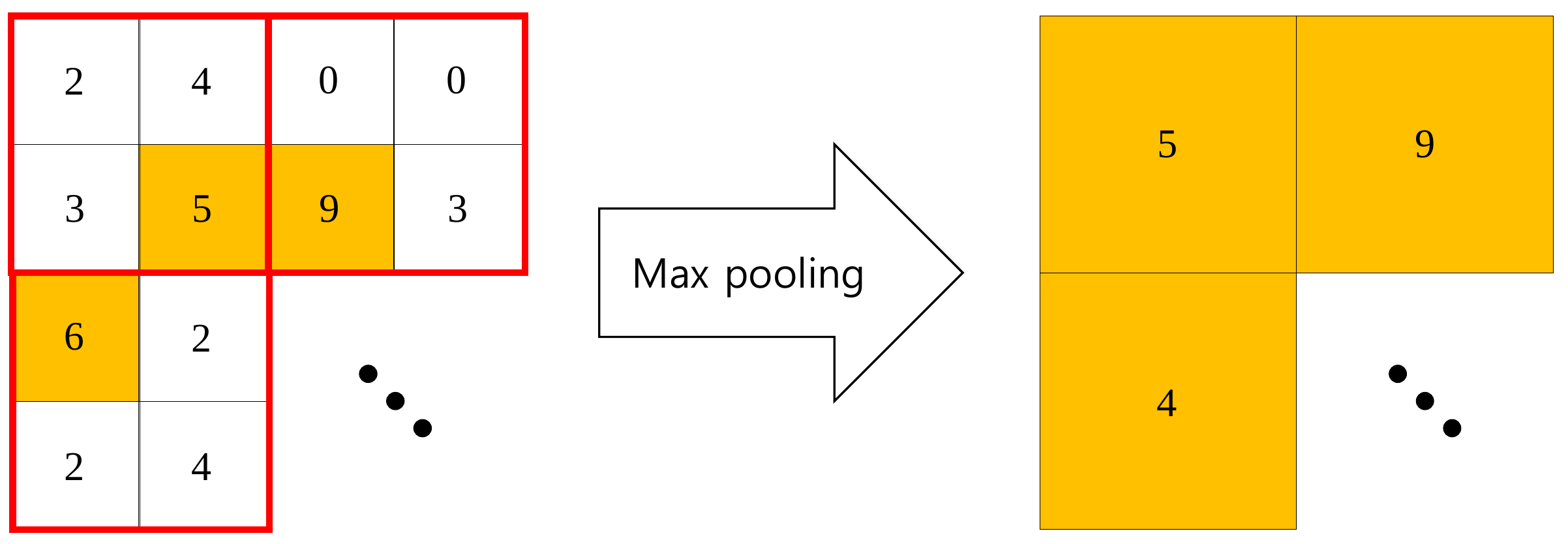}}
\caption{Max pooling takes the maximum value within the pooling block. The orange areas denote the locations from which the output values are taken.}
\label{fig2}
\end{figure}

Max pooling computes the output using the maximum function. The output from the max pooling is defined as 
\begin{equation}
{o_{0,0}} = \max ({f_{0,0}},{f_{0,1}},{f_{1,0}},{f_{1,1}})
\label{eq3}
\end{equation}
and its example is given in Figure \ref{fig2}.

\subsection{Average pooling}
\label{3B}
Average pooling returns the average of each pooling block in the feature map. The averaging function is given by
\begin{equation}
{o_{0,0}} = \frac{1}{4}({f_{0,0}} + {f_{0,1}} + {f_{1,0}} + {f_{1,1}}).
\label{eq4}
\end{equation}

\begin{figure}[htbp]
\centerline{\includegraphics[width=0.9\linewidth]{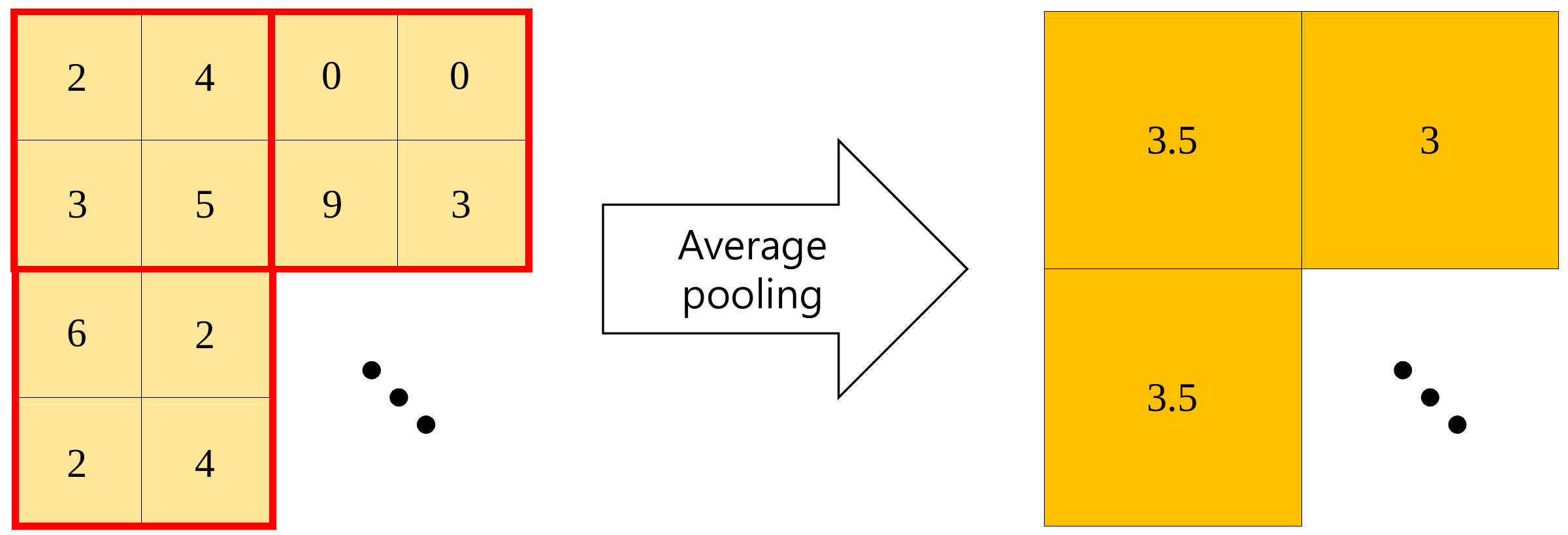}}
\caption{Average pooling averages the feature-map entries in each pooling block.}
\label{fig3}
\end{figure}

In average pooling, all inputs of the corresponding block in the feature map equally contribute to the output. A simple example of average pooling is given in Figure \ref{fig3}.

\subsection{Stride pooling (stride convolution)}
\label{3C}
\begin{figure}[htbp]
\centerline{\includegraphics[width=0.9\linewidth]{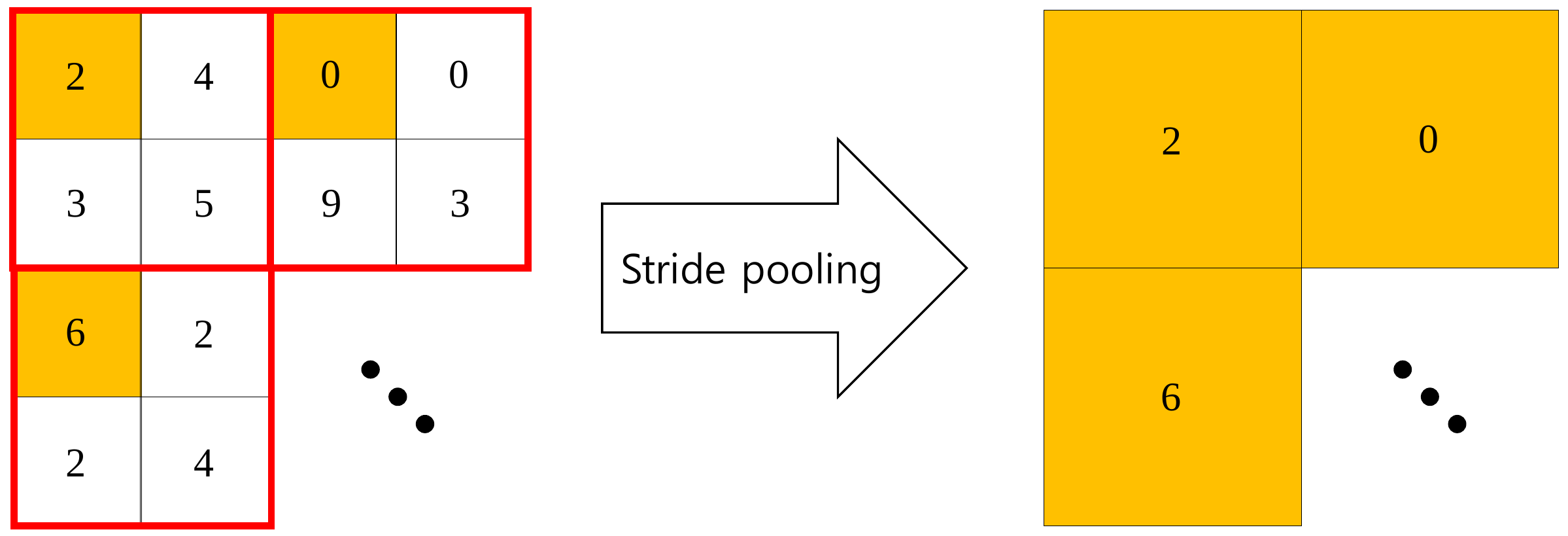}}
\caption{Stride convolution can be considered as stride pooling without the intermediate operations. Stride pooling takes the values from certain regions of the input.}
\label{fig4}
\end{figure}

Stride convolution reduces the size of the feature map by applying the stride in the convolution layer, without invoking an extra function. The convolution layer with a stride performs the same function as the convolution layer without a stride, followed by fixed-position pooling. The stride pooling is represented by 
\begin{equation}
{o_{0,0}} = {f_{0,0}} ,
\label{eq5}
\end{equation}
and is demonstrated by a simple example in Figure \ref{fig4}.

\section{Universal Pooling}
\label{4}
This section proposes our new pooling method named universal pooling, which automatically determines the pooling mechanism by training the pooling weights. Universal pooling can be considered as a channel-wise local spatial attention module. Here we show that the proposed method encompasses max pooling, average pooling and stride pooling. For simplicity, we assume a pooling size of 2 and a stride of 2, but the concept is easily extended to more general cases.

\subsection{Motivation}
\label{4A}
\begin{figure}[htbp]
\centerline{\includegraphics[width=0.9\linewidth]{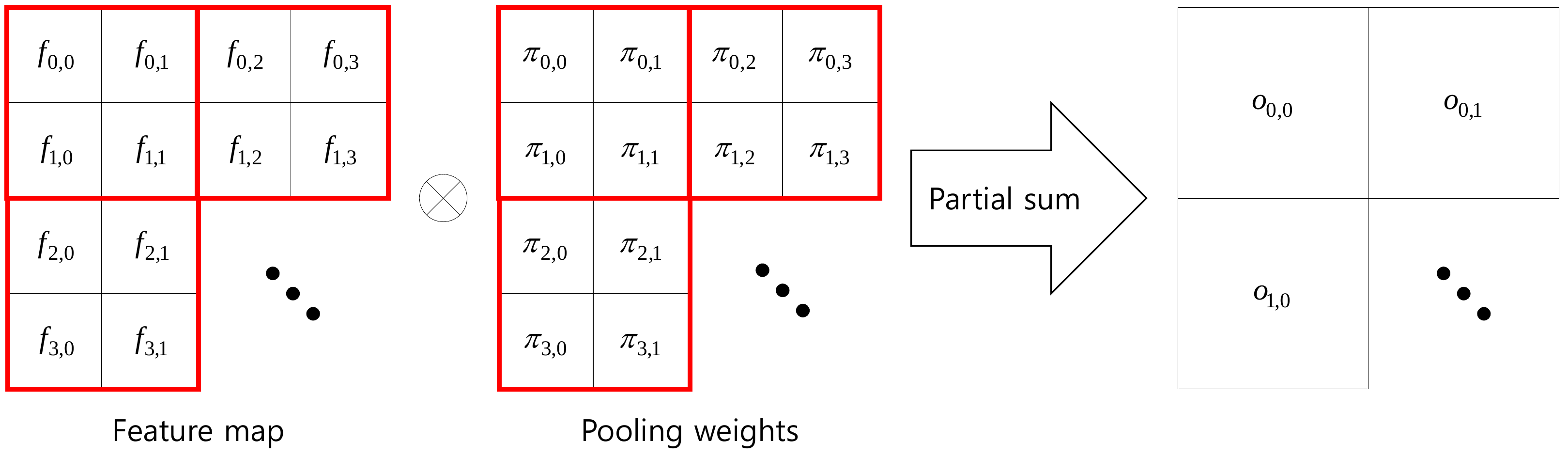}}
\caption{Standard pooling can be considered as a linear combination within each pooling block. The weights in this linear combination are called the pooling weights.}
\label{fig5}
\end{figure}

The popular standard pooling methods, demonstrated by simple examples in Section III, are represented by linear combinations of the weighted features within a pooling block in the feature map. More specifically, the simple poolings are expressed as 

\begin{align}
{o_{0,0}} & = {\mathrm{universal\_pooling}} \left( {f_{0,0}},{f_{0,1}},{f_{1,0}},{f_{1,1}} \right) \nonumber \\
 & = {\pi _{0,0}}{f_{0,0}} + {\pi _{0,1}}{f_{0,1}} + {\pi _{1,0}}{f_{1,0}} + {\pi _{1,1}}{f_{1,1}} \label{eq6}\\
 & \buildrel \Delta \over = {\boldsymbol{\pi }} \otimes {\boldsymbol{f}} \nonumber
\end{align}
where ${\boldsymbol{\pi }}$ denote the pooling weights, ${\boldsymbol{f}}$ is the input feature map, and the operator $\otimes$ performs linear combinations within the pooling block. Equation \eqref{eq6} interprets the pooling operation as an element-wise multiplication and summation within each pooling-size block in the feature map, as shown in Figure \ref{fig5}. In universal pooling, the pooling weight ${\boldsymbol{\pi }}$ is not manually selected, but is automatically determined from the input image or feature map. The pooling weight is also trained by back-propagation, by a mechanism similar to an attention map.

\subsection{Structure}
\label{4B}
The pooling operation computes a scalar value representing the pooling block in the feature map. In implementations, the universal pooling operation should be divided into two smaller operations. The first sub-operation determines the contribution of each element in each pooling block of the feature map, which is equivalent to determining the pooling weight ${\boldsymbol{\pi }}$. The second sub-operation applies the $\otimes$ operator to the pooling weights and feature map, and returns the representative value of an element in the next-layer feature map.

The characteristics of the proposed universal pooling are similar to those of the existing pooling methods, and are subjected to the following constraints: (C1) Pooling is conducted in a channel-wise manner and excludes the information from all other channels. The pooling weight ${\boldsymbol{\pi }}$ is thus trained separately from channel to channel. (C2) Only the feature map elements in a given pooling block are pooled; the elements outside of the block are excluded. When training the pooling weights, the receptive fields of the weights are thus restricted to their corresponding pooling block. (C3) The scale of the features is unchanged by the pooling. For this purpose, the pooling weights are normalized such that ${\pi _{0,0}} + {\pi _{0,1}} + {\pi _{1,0}} + {\pi _{1,1}} = 1$. Our universal pooling is designed to satisfy constraints C1, C2, and C3. As shown in Figure 6, the universal pooling module has two main blocks: Block B1, in which the pooling weights are computed, and Block B2, in which the pooling is performed using the pooling weights.

\begin{figure}[htbp]
\centerline{\includegraphics[width=0.9\linewidth]{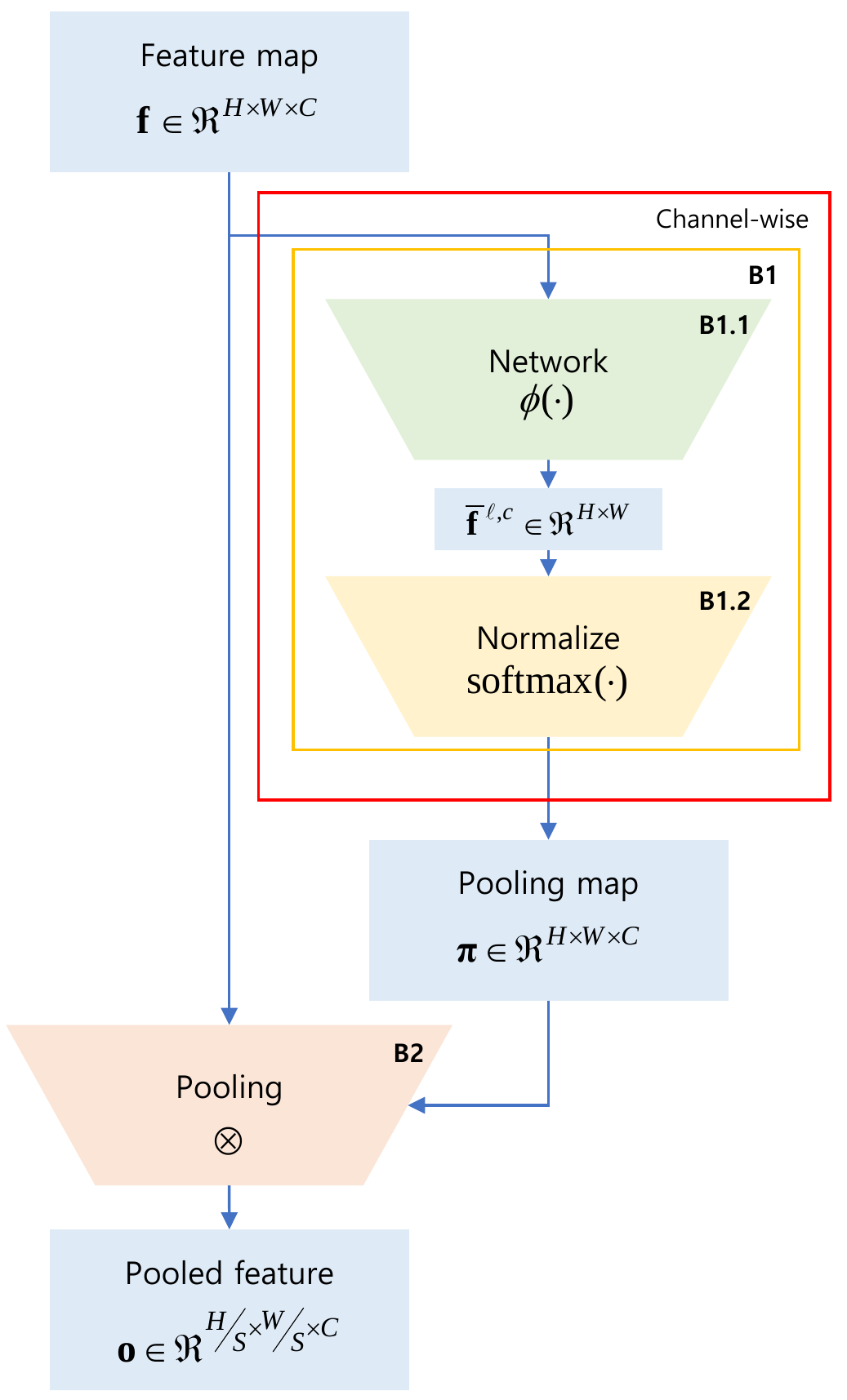}}
\caption{Structure of the proposed universal pooling module. The universal pooling module has two main blocks. B1 computes the pooling weight, and B2 performs the pooling using the pooling weights.}
\label{fig6}
\end{figure}

The pooling-weight block consists of several layers-the convolution, fully connected, batch normalization, and activation layers-as in other CNN modules. Using the ${\mathrm{softmax}} \left( \cdot \right)$ function, the output of this block is then normalized over the corresponding pooling block to obtain ${\pi _{0,0}} + {\pi _{0,1}} + {\pi _{1,0}} + {\pi _{1,1}} = 1$. The second block applies the $\otimes$ operator to the pooling weights and feature map over the pooling block to obtain the pooling result. To simplify the description, we assume identical size and stride number in the pooling method of the proposed approach, but unequal size and stride number are equally valid.  

\begin{figure}[htbp]
\centerline{\includegraphics[width=0.9\linewidth]{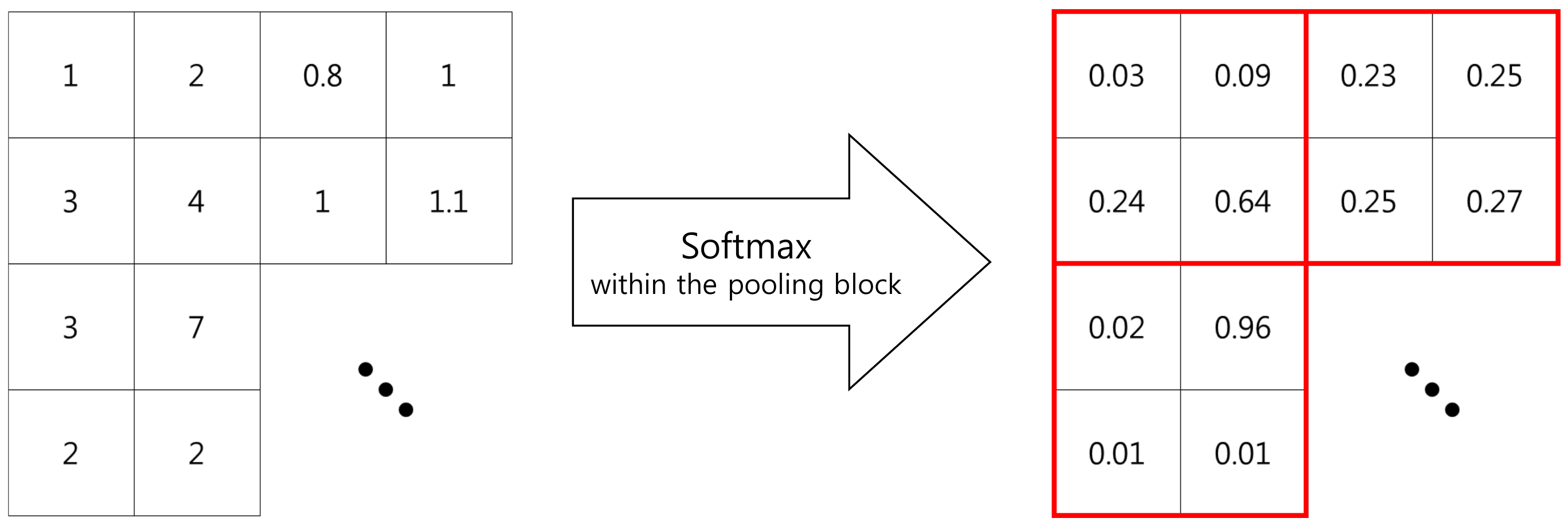}}
\caption{Application of softmax within the pooling block. Each red square delineates a pooling block. Note that the pooling weights within each pooling block sum to one.}
\label{fig7}
\end{figure}

To clarify the relationship between the input and output of universal pooling, we give the equations of the proposed pooling. Before normalization, the output of the first block (B1) is represented as
\begin{equation}
\begin{array}{l}
{{{\boldsymbol{\overline{f}}}}^{\ell ,c}} = \phi ({{\boldsymbol{f}}^{\ell ,c}};{\boldsymbol{W}})\\
{{\boldsymbol{f}}^{\ell ,c}} = \left( {f_{i,j}^{\ell ,c}} \right) \in {\Re ^{H \times W}},{{{\boldsymbol{\overline{f}}}}^{\ell ,c}} = \left( {\overline{f}_{i,j}^{\ell ,c}} \right) \in {\Re ^{H \times W}}
\end{array}
\label{eq7}
\end{equation}
where $\phi ( \cdot )$ denotes the CNN module receiving ${\boldsymbol{f}}^{\ell,c}$, the feature map of channel $c$ in the $l$-th layer, and returning the temporary feature map ${{\boldsymbol{\overline{f}}}^{\ell ,c}} = \phi ({\boldsymbol{f}}^{\ell ,c};{\boldsymbol{W}})$ of channel $c$ (which is the same size as ${\boldsymbol{f}}^{\ell,c}$). ${\boldsymbol{W}}$ is a trainable parameter inside the network $\phi ( \cdot )$. To satisfy constraint (C3), the corresponding pooling block is processed by the softmax function, which guarantees that the pooling weights within the pooling block sum to one, as shown in Figure \ref{fig7}. This operation is represented by
\begin{equation}
\begin{gathered}
  {{\boldsymbol{\pi }}^{\ell ,c}} = {\mathrm{softmax}}({{{\boldsymbol{\overline{f}}}}^{\ell ,c}}) \hfill \\
  \pi _{pS + m,qS + n}^{\ell ,c} \hfill \\
  \quad = \frac{1}{{\sum\limits_{i = 0}^{S - 1} {\sum\limits_{j = 0}^{S - 1} {\exp (\overline{f}_{pS + i,qS + j}^{\ell ,c})} } }}\exp (\overline{f}_{pS + m,qS + n}^{\ell ,c}) \hfill \\
  {{\boldsymbol{\pi }}^{\ell ,c}} = \left( {\pi _{i,j}^{\ell ,c}} \right) \in {\Re ^{H \times W}} \hfill\\
  p = 0,1, \cdots ,\left\lfloor {{\raise0.7ex\hbox{$H$} \!\mathord{\left/
 {\vphantom {H S}}\right.\kern-\nulldelimiterspace}
\!\lower0.7ex\hbox{$S$}}} \right\rfloor  - 1,q = 0,1, \cdots ,\left\lfloor {{\raise0.7ex\hbox{$W$} \!\mathord{\left/
 {\vphantom {W S}}\right.\kern-\nulldelimiterspace}
\!\lower0.7ex\hbox{$S$}}} \right\rfloor  - 1 \hfill \\ 
\end{gathered} 
\label{eq8}
\end{equation}
where $S$ denotes the stride number and size of the pooling, ${\mathrm{softmax}} \left( \cdot \right)$ is applied to an $S \times S$ pooling block, and  denotes the learnable pooling weight. Here, ${{\boldsymbol{\pi }}^{\ell ,c}}$ and $\overline{f}_{i,j}$ are the same size.

Then, the output of the first block (B1) is related to its input as follows:
\begin{equation}
{{\boldsymbol{\pi }}^{\ell ,c}} = {\mathrm{softmax}}(\phi ({{\boldsymbol{f}}^{\ell ,c}};{\boldsymbol{W}})).
\label{eq9}
\end{equation}

In \eqref{eq9}, the learnable parameter $\boldsymbol{W}$ inside the module determines the pooling function. To satisfy constraint (C1), $\phi ( \cdot )$ is applied only to the corresponding channel. The universal pooling then learns the different pooling functions in the different channels.
In the second block (B2), pooling weights are assigned to the feature map, and pooling is performed through the $ \otimes $ operation. The output from (B2) is represented by 
\begin{equation}
\begin{gathered}
  {{\boldsymbol{o}}^{\ell ,c}} = {{\boldsymbol{\pi }}^{\ell ,c}} \otimes {{\boldsymbol{f}}^{\ell ,c}} \hfill \\
  o_{p,q}^{\ell ,c} = \sum\limits_{m = 0}^{S - 1} {\sum\limits_{n = 0}^{S - 1} {\pi _{pS + m,qS + n}^{\ell ,c} \cdot f_{pS + m,qS + n}^{\ell ,c}} }  \hfill \\
  {{\boldsymbol{o}}^{\ell ,c}} = \left( {o_{p,q}^{\ell ,c}} \right) \in {\Re ^{\scriptsize{\left\lfloor {{\raise0.7ex\hbox{$H$} \!\mathord{\left/
 {\vphantom {H S}}\right.\kern-\nulldelimiterspace}
\!\lower0.7ex\hbox{$S$}}} \right\rfloor  \times \left\lfloor {{\raise0.7ex\hbox{$W$} \!\mathord{\left/
 {\vphantom {W S}}\right.\kern-\nulldelimiterspace}
\!\lower0.7ex\hbox{$S$}}} \right\rfloor }}} \hfill \\
  p = 0,1, \cdots ,\left\lfloor {{\raise0.7ex\hbox{$H$} \!\mathord{\left/
 {\vphantom {H S}}\right.\kern-\nulldelimiterspace}
\!\lower0.7ex\hbox{$S$}}} \right\rfloor  - 1,{\text{ }}q = 0,1, \cdots ,\left\lfloor {{\raise0.7ex\hbox{$W$} \!\mathord{\left/
 {\vphantom {W S}}\right.\kern-\nulldelimiterspace}
\!\lower0.7ex\hbox{$S$}}} \right\rfloor  - 1{\text{ }} \hfill \\ 
\end{gathered}
\label{eq10}
\end{equation}
where the output ${{\boldsymbol{o}}^{\ell ,c}}$ from (B2) defines a new feature map in the $\left(\ell+1\right)$-th layer. Obviously, the scale of the features is unchanged and constraint (C2) is satisfied because the pooling weights are already normalized by \eqref{eq9}. In \eqref{eq10}, the pooling weights   can be rewritten as
\begin{equation}
{{\boldsymbol{o}}^{\ell ,c}} = {{\boldsymbol{\pi }}^{\ell ,c}} \otimes {{\boldsymbol{f}}^{\ell ,c}} = {\mathrm{softmax}}(\phi ({{\boldsymbol{f}}^{\ell ,c}};{\boldsymbol{W}})) \otimes {{\boldsymbol{f}}^{\ell ,c}}.
\label{eq11}
\end{equation}
Because all functions in this equation are simple multiplications and additions, all the learnable parameters are easily updated by back-propagation.

\subsection{Relationship between Universal Pooling and the Existing Pooling Methods}
\label{4C}
As mentioned above, the proposed pooling module can be considered as a channel-wise local spatial attention module. This subsection demonstrates that the popular existing pooling methods are special cases of universal pooling.

\subsubsection{Average Pooling}
\label{4CA}
Average pooling averages the elements inside the pooling block of the feature map. The averaging is performed as follows:
\begin{equation}
o_{p,q}^{\ell ,c} = \frac{1}{{{S^2}}}\sum\limits_{m = 0}^{S - 1} {\sum\limits_{n = 0}^{S - 1} {f_{pS + m,qS + n}^{\ell ,c}} } 
\label{eq12}
\end{equation}
where the pooling block is an $S \times S$ matrix. Using the pooling weights ${{\boldsymbol{\pi }}^{\ell ,c}}$, we can rewrite \eqref{eq12} as
\begin{equation}
\begin{gathered}
  o_{p,q}^{\ell ,c} = \sum\limits_{m = 0}^{S - 1} {\sum\limits_{n = 0}^{S - 1} {\pi _{pS + m,qS + n}^{\ell ,c} \cdot f_{pS + m,qS + n}^{\ell ,c}} } , \hfill \\
  \pi _{pS + m,qS + n}^{\ell ,c} = \frac{1}{{{S^2}}} . \hfill \\ 
\end{gathered} 
\label{eq13}
\end{equation}
The proposed universal pooling behaves like average pooling if all pooling weights are equal and sum to unity. The pooling weights are constant for any input image. This implies that before ${\mathrm{softmax}}( \cdot )$ the function, the output $\overline{f}^{\ell ,c}$ of (B1.1) is the same over the whole pooling block. Thus, when universal pooling performs like average pooling, its form reduces to 
\begin{equation}
\begin{gathered}
  o_{p,q}^{\ell ,c} = \sum\limits_{m = 0}^{S - 1} {\sum\limits_{n = 0}^{S - 1} {\pi _{pS + m,qS + n}^{\ell ,c} \cdot f_{pS + m,qS + n}^{\ell ,c}} } , \hfill \\
  \pi _{pS + m,qS + n}^{\ell ,c} \hfill\\
  \quad = \frac{1}{{\sum\limits_{i = 0}^{S - 1} {\sum\limits_{j = 0}^{S - 1} {\exp (\overline{f}_{pS + i,qS + j}^{\ell ,c})} } }}\exp (\overline{f}_{pS + m,qS + n}^{\ell ,c}) \hfill\\
  \quad = \frac{1}{{{S^2}}} , \hfill \\
  \overline{f}_{pS + i,qS + j}^{\ell ,c} = \overline{f}_{pS + x,qS + y}^{\ell ,c}{\text{        }}(x \ne i,y \ne j) . \hfill \\ 
\end{gathered} 
\label{eq14}
\end{equation}
In the simplest example of average pooling by the proposed universal pooling, the learnable parameter $\boldsymbol{W}$ is set to 0. In this case, all outputs are zero and the ${\mathrm{softmax}}( \cdot )$ function generates equal pooling weights that sum to one.

\subsubsection{Max Pooling}
\label{4CB}
Max pooling outputs the maximum value of the elements inside a given pooling block. The max pooling function can be written as follows.
\begin{equation}
o_{p,q}^{\ell ,c} = \mathop {\max }\limits_{0 \leqslant m,n \leqslant S - 1} f_{pS + m,qS + n}^{\ell ,c} .
\label{eq15}
\end{equation}

In terms of the pooling weights ${{\boldsymbol{\pi }}^{\ell ,c}}$, \eqref{eq15} can be rewritten as
\begin{equation}
\begin{gathered}
  o_{p,q}^{\ell ,c} = \sum\limits_{m = 0}^{S - 1} {\sum\limits_{n = 0}^{S - 1} {\pi _{pS + m,qS + n}^{\ell ,c} \cdot f_{pS + m,qS + n}^{\ell ,c}} } , \hfill \\
  \pi _{pS + m,qS + n}^{\ell ,c} = \left\{ {\begin{array}{*{20}{c}}
  1&{m,n = \mathop {\arg \max }\limits_{0 \leqslant i,j \leqslant S - 1} f_{pS + i,qS + j}^{l,c}} \\ 
  0&{otherwise} 
\end{array}} . \right. \hfill \\ 
\end{gathered}
\label{eq16}
\end{equation}

The pooling weight is set to one at the site containing the maximum value in the block, and zero at all other sites. If the output $\overline{f}^{\ell ,c}$ of (B1.1) equals the input feature map $f^{\ell ,c}$ before invoking the ${\mathrm{softmax}}( \cdot )$ function, the universal pooling becomes
\begin{equation}
\begin{gathered}
  o_{p,q}^{\ell ,c} = \sum\limits_{m = 0}^{S - 1} {\sum\limits_{n = 0}^{S - 1} {\pi _{pS + m,qS + n}^{\ell ,c} \cdot f_{pS + m,qS + n}^{\ell ,c}} } , \hfill \\
  \pi _{pS + m,qS + n}^{\ell ,c} = {\mathrm{softmax}}({{{\boldsymbol{\overline{f}}}}^{\ell ,c}}) = {\mathrm{softmax}}({{\boldsymbol{f}}^{\ell ,c}}) \hfill \\
  \qquad \qquad \; \approx \left\{ {\begin{array}{*{20}{c}}
  1&{m,n = \mathop {\arg \max }\limits_{0 \leqslant i,j \leqslant S - 1} f_{pS + i,qS + j}^{\ell,c}} \\ 
  0&{otherwise} 
\end{array}} .\right. \hfill \\ 
\end{gathered} 
\label{eq17}
\end{equation}

In this case, the proposed universal pooling becomes very similar to max pooling. In summary, the condition under which the proposed method behaves like max pooling is
\begin{equation}
f^{\ell ,c} = \overline{f}^{\ell ,c} = \phi (f^{\ell ,c}) .
\label{eq18}
\end{equation}
That is, universal pooling reduces to max pooling when (B1.1) is trained as an identity mapping.

\subsubsection{Stride Pooling}
\label{4CC}
Stride pooling obtains an output from a specific location inside an area. When the specific location is the top-left site of the pooling block, the stride pooling is described by
\begin{equation}
o_{p,q}^{\ell ,c} = f_{pS,qS}^{\ell ,c}.
\label{eq19}
\end{equation}

In terms of the pooling weights, \eqref{eq19} is rewritten as
\begin{equation}
\begin{gathered}
  o_{p,q}^{\ell ,c} = \sum\limits_{m = 0}^{S - 1} {\sum\limits_{n = 0}^{S - 1} {\pi _{pS + m,qS + n}^{\ell ,c} \cdot f_{pS + m,qS + n}^{\ell ,c}} } , \hfill \\
  \pi _{pS + m,qS + n}^{\ell ,c} = \left\{ {\begin{array}{*{20}{c}}
  1&{m,n = 0,0} \\ 
  0&{otherwise} 
\end{array}} . \right. \hfill \\ 
\end{gathered}
\label{eq20}
\end{equation}

The proposed universal pooling behaves like stride pooling when the first block (B1) outputs unity at the top-left location of the pooling block, and zero at the remaining positions. 

Above, we showed that the proposed universal pooling includes the popular existing pooling methods as special cases. This implies that the performance of the existing pooling methods can be improved by replacing them with universal pooling. In particular, by appropriately training $\boldsymbol{W}$ in universal pooling, we can expect to improve the pooling to an unprecedented level, with concomitant improvement in CNN performance.

\section{Experimentation}
\label{5}
In this section, the proposed method is experimentally tested on two benchmark classification datasets: CIFAR10 \cite{b17} and Places2 \cite{b18}. The VGG19 and ResNet18 datasets are employed as backbone networks.

\subsection{Implementation}
\label{5A}
\begin{figure*}[!t]
    \centering
    \subfloat[]{
    \raisebox{11mm}{
        \includegraphics[width=0.25\textwidth]{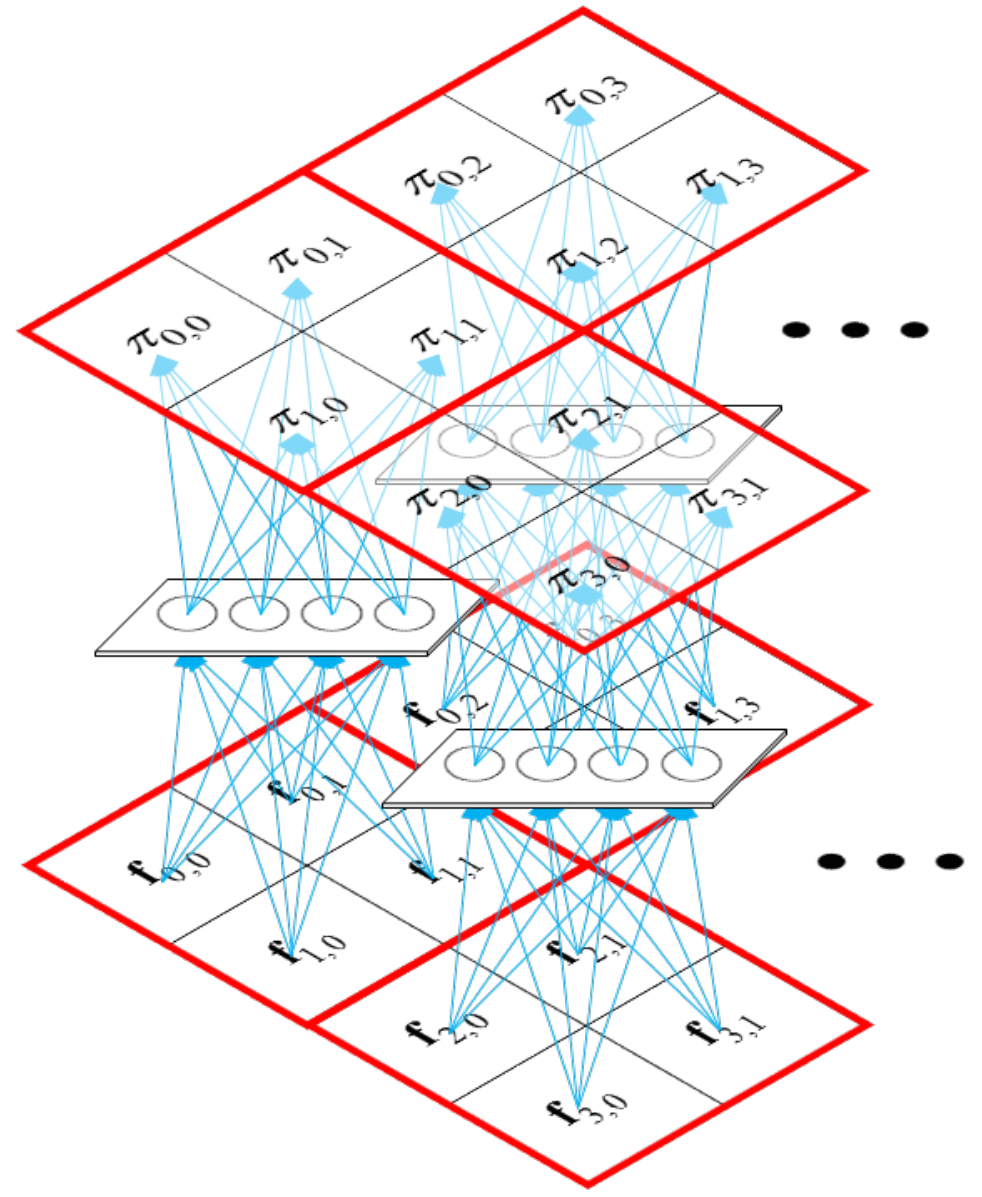}
        \label{fig8a}
    }}
    \hfill
    \subfloat[]{
    \raisebox{11mm}{
        \includegraphics[width=0.25\textwidth]{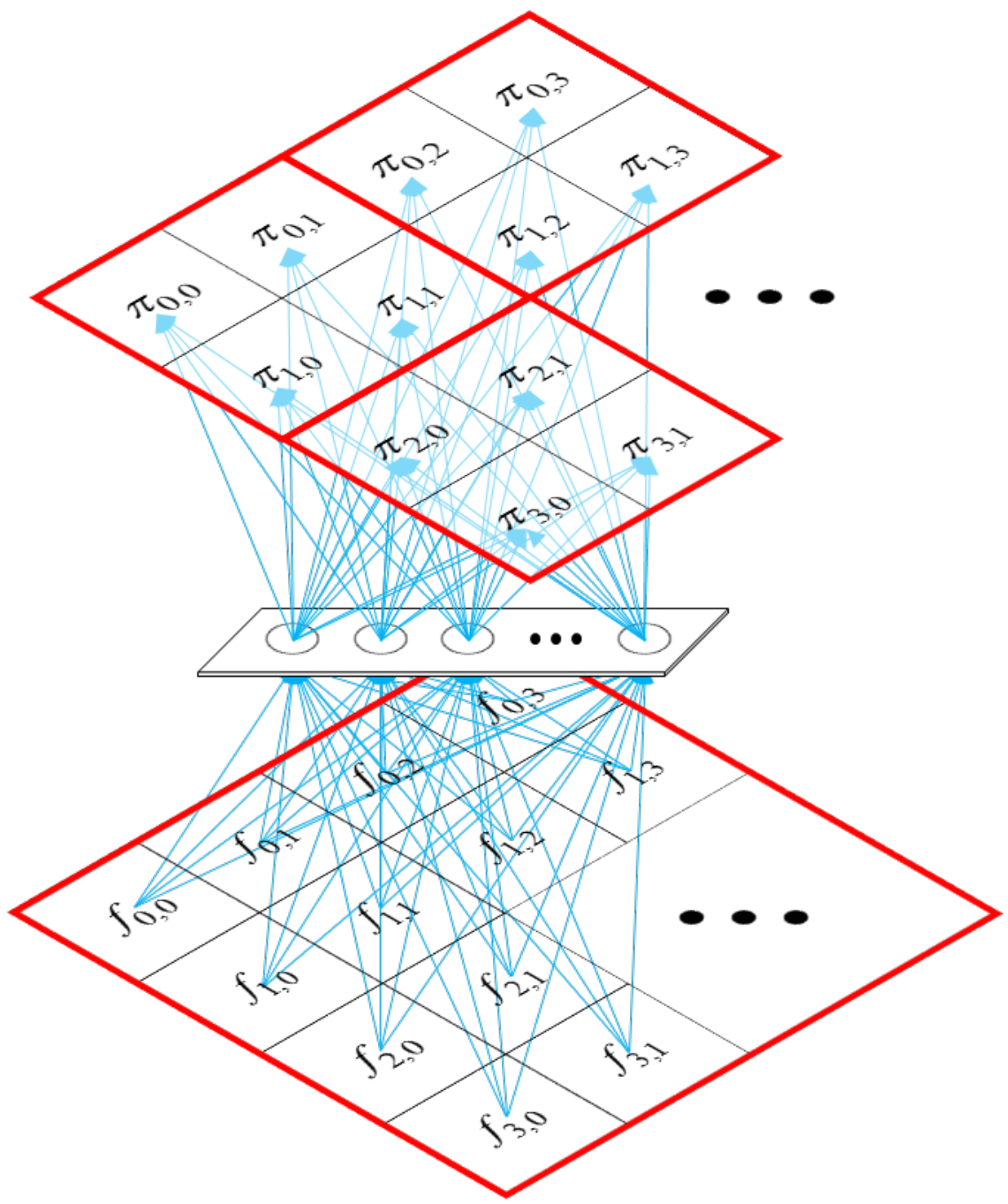}
        \label{fig8b}
    }}
    \hfill
    \subfloat[]{
        \includegraphics[width=0.28\textwidth]{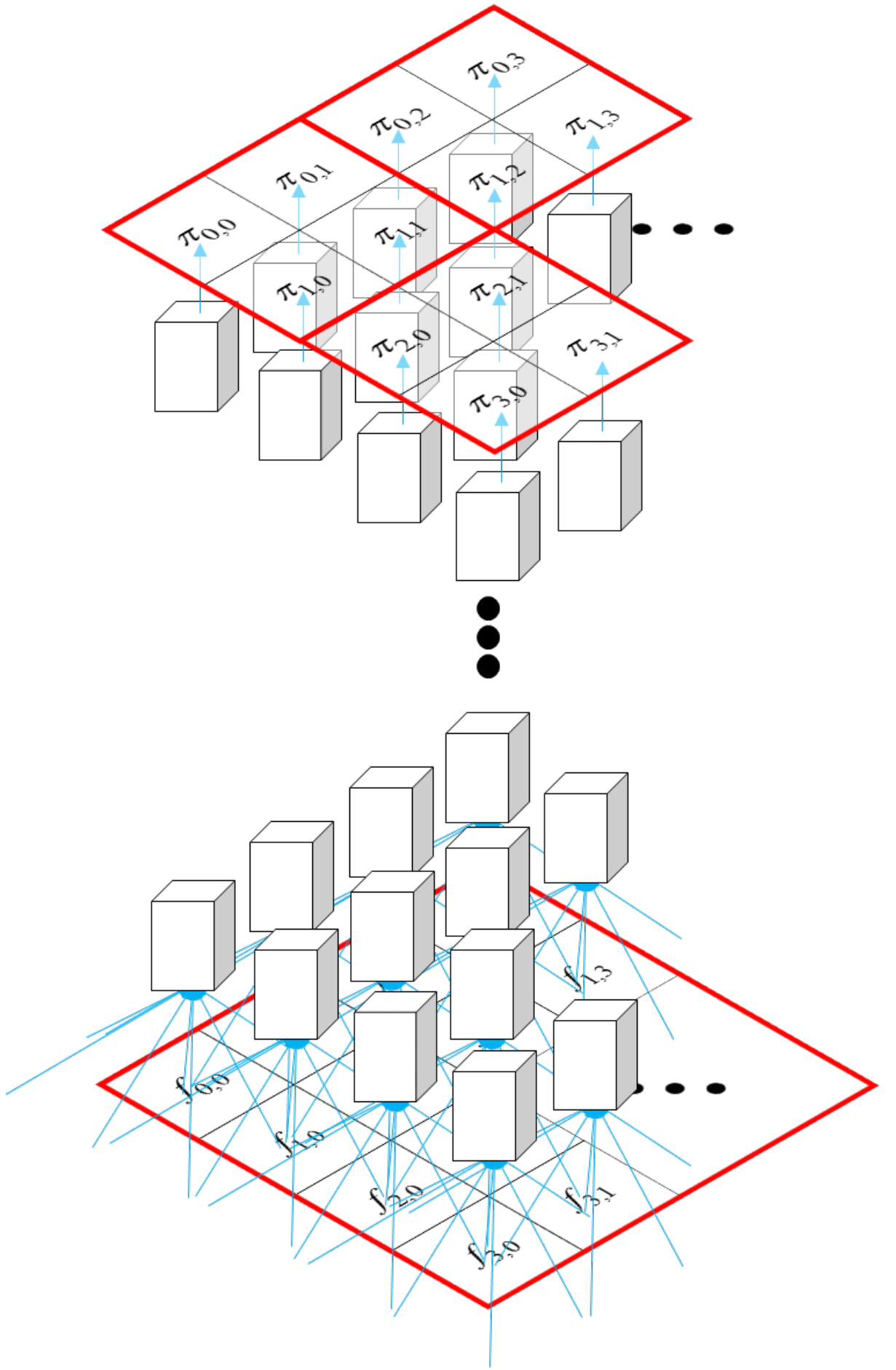}
        \label{fig8c}
    }
    \caption{Local and global pooling implemented by fully connected and convolutional layers. The red squares delineate the pooling blocks. Local pooling shares the weights in the fully connected layers of the same channel. (a) Local pooling with a pooling module of fully connected layers; (b) Global pooling with a pooling module of fully connected layers; (c) Global pooling with a pooling module of several convolution layers.}
    \label{fig8}
\end{figure*}

Pooling can be divided into local pooling and global pooling. In local pooling, the pooling block is smaller than the feature map, whereas in global pooling, it covers the entire feature map. In the present experiment, local pooling was implemented by one or two fully connected layers, as shown in Figure \ref{fig8a}. Global pooling also can be implemented either by fully connected layers (Figure \ref{fig8b}), or by several convolution layers (Figure \ref{fig8c}).

\subsection{CIFAR10}
\label{5B}
\begin{figure}[htbp]
\centerline{\includegraphics[width=0.9\linewidth]{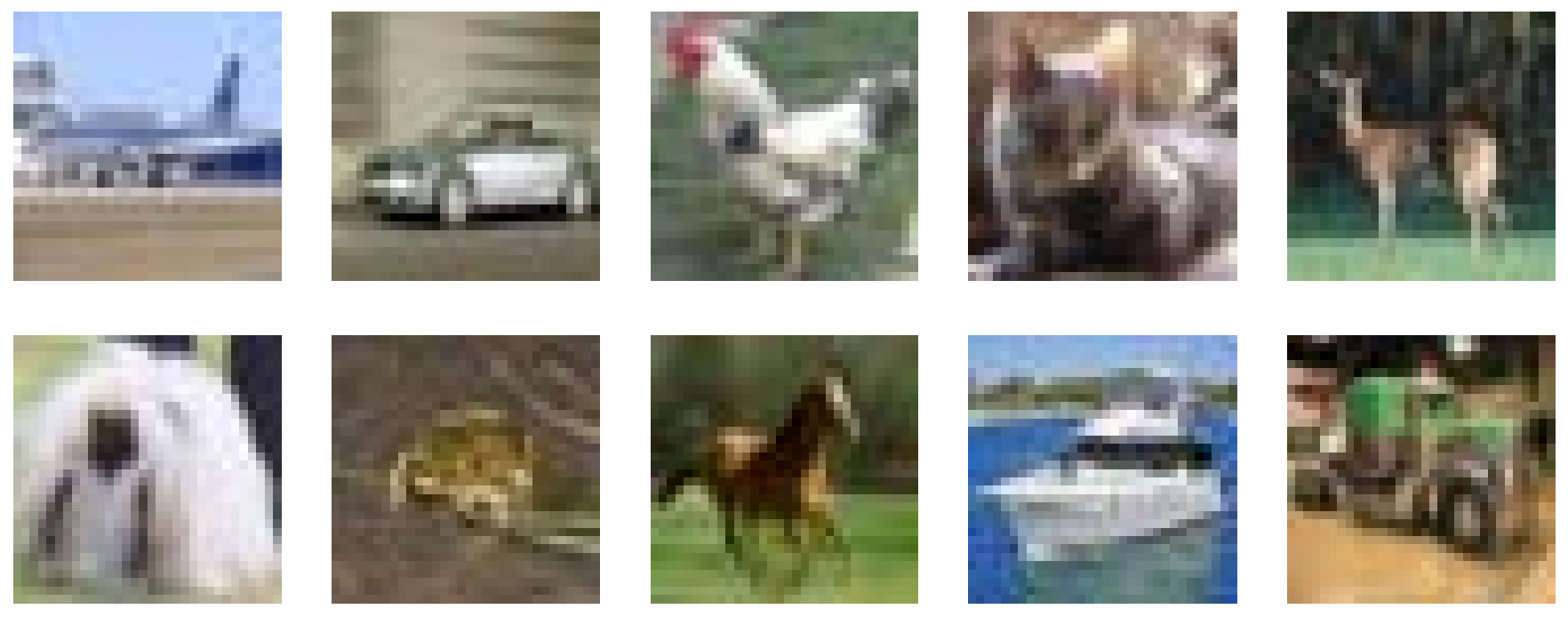}}
\caption{Examples from the CIFAR10 dataset, which comprises small $32 \times 32$ RGB images.}
\label{fig9}
\end{figure}

The Canadian Institute for Advanced Research 10 (CIFAR10) dataset is a classification dataset containing 60,000 images \cite{b17}, each of size $32 \times 32$ on the RGB channels. The dataset contains ten object classes: airplane, automobile, bird, cat, deer, dog, frog, horse, ship, and truck. Typical images contained in the dataset are shown in Figure \ref{fig9}. The CNNs were trained by the stochastic gradient descent (SGD) method. The initial learning rate, momentum, and weight decay in the SGD were set to 0.1, 0.9, and 0.0001, respectively. The whole training procedure was implemented over 450 epochs, reducing the learning rate by one tenth at 150-epoch intervals.


The backbone networks were two pre-trained models: VGG19 and ResNet18. The original VGG19 network consists of 16 convolution layers, 3 fully connected layers, and 5 max pooling layers. Instead of global pooling, VGG vectorizes the last feature map and passes it to the fully connected layer. The existing VGG structure, which is designed for large-sized images, was here modified for processing small images by employing one fully connected layer. As the last feature map is already vectorized by the final pooling process, it requires no further vectorization, so the final pooling can be considered as a global pooling. This structure of the network is given in Table \ref{tab1}.

The original VGG network (here termed V1) performs five max poolings. When the input image is a $32 \times 32$ CIFAR 10 image, the last-pooled map and the input feature map are the same size. Thus, the final pooling is considered as a global pooling while the preceding poolings are local poolings. As competing methods for comparison, we also employed five variants of VGG19. In variant V2, all five max poolings were replaced with their average pooling procedures. In variant V3, the local max pooling was replaced with local stride pooling in the first four pooling layers, and GAP was applied in the last pooling layer. The structure of V3 matched that of ResNet. Variants V4-V6 were constructed by various combinations of max and average pooling \cite{b15}, as shown in Table \ref{tab3}. In this table, "Mixed" refers to a linear combination of max and average pooling with real-numbered weights. Meanwhile, "Gated-channel" and "Gated-layer" refer to gated max-average pooling in which the gating mask is a linear combination of the weighted channels and weighted layers, respectively. Four variants of the proposed method, labeled P1-P5, were also constructed for a comparison evaluation. In variants P1-P3, both local and global poolings were implemented by fully connected layers, while in variants P4 and P5, the first four local poolings were implemented by fully connected layers, and the final (global) pooling was implemented by a convolution layer (see Table \ref{tab3}). Interestingly, the average pooling V2 (rather than the original VGG V1) delivered the best performance among the standard pooling methods. The combined max and average pooling methods (V4-V6) outperformed the standard pooling methods (V1-V3). Meanwhile, among variants P1-P5 of the proposed universal pooling method, constructed as competitors against the standard methods, P4 and P3 delivered the best and second-best performances, respectively. Overall, increasing the number of layers improved the network performance. However, the performance was degraded in the deepest network (P5), possibly because the network overfitted the training data.

\begin{figure*}[!t]
    \centering
    \subfloat[]{
        \includegraphics[width=0.45\textwidth]{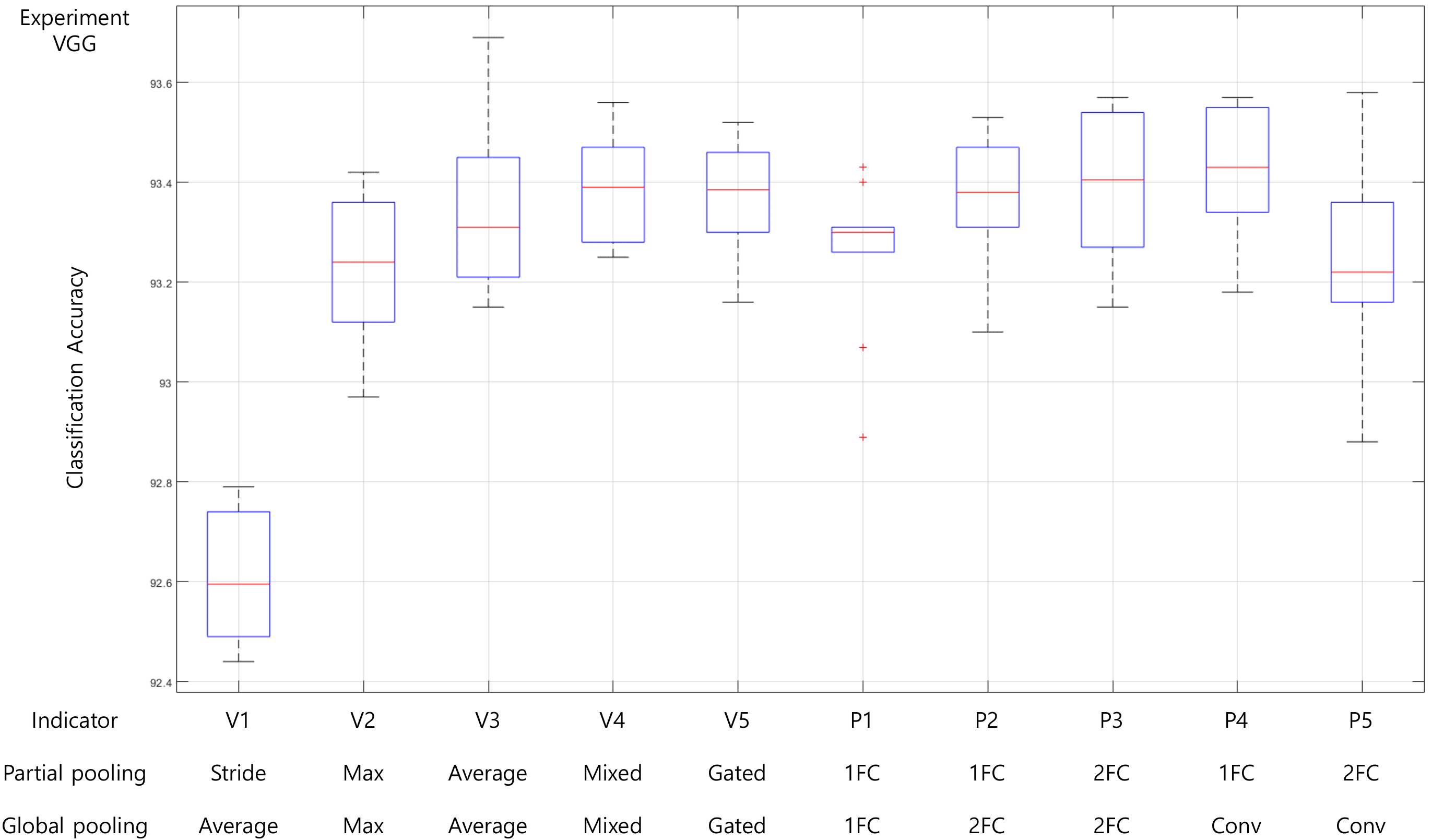}
        \label{fig10a}
    }
    \quad\quad\quad
    \subfloat[]{
        \includegraphics[width=0.45\textwidth]{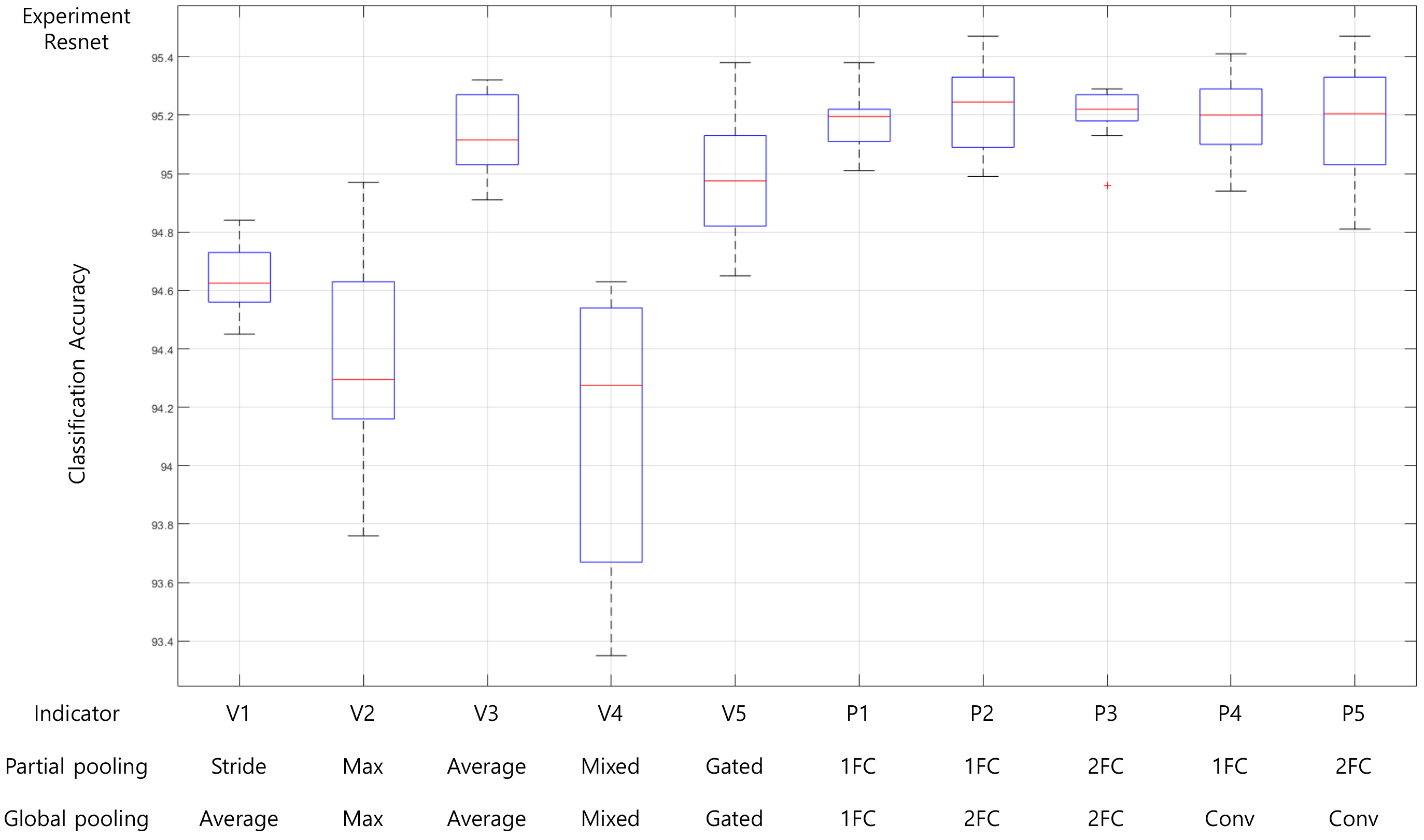}
        \label{fig10b}
    }
    \caption{Boxplot of experiments on the CIFAR10 dataset, performed in the (a)VGG architecture, and (b)the ResNet architecture. Red lines denote the experimental averages. The proposed method outperformed the other pooling methods.}
    \label{fig10}
\end{figure*}

Second, let us consider ResNet. The original ResNet18 consists of 17 convolution layers, 1 fully connected layer, 5 local pooling layers, and 1 global pooling layer. In the original ResNet, the local and global poolings are performed by striding and averaging (V3), respectively. The existing ResNet structure, which is suitable for large-sized images, was here modified for small-image processing by employing three (rather than five) local pooling layers. The network structure is given in Table \ref{tab2}. As in VGG, two standard pooling methods (V1 with max pooling and V2 with average pooling) and three combinations of max and average poolings (V4-V6) were constructed as competing methods. All experiments were conducted ten times, and the performances are summarized in Table \ref{tab2} and Figure \ref{fig10}. As confirmed in the table, average pooling delivered the best performance among the standard pooling methods (V1-V3, where V3 is the original ResNet18). The performance of average pooling (V2) was significantly higher than that of max pooling (V1). Interestingly, the average pooling outperformed even the combination methods (V4-V6). This suggests that training all poolings to behave like average pooling is a difficult task. However, the proposed universal pooling outperformed the average pooling, as shown in Table \ref{tab2}. The overall best performance was achieved by a network with one fully connected layer for local pooling, and two fully connected layers for global pooling. As demonstrated in this experiment, the optimal performance is dictated by the structure of the entire network, and the proposed universal pooling outperformed all standard pooling methods and their combinations proposed in \cite{b15}.

\begin{table}
\centering
\caption{VGG network, modified to suit the CIFAR10 dataset.}
\setlength{\tabcolsep}{3pt}
\begin{tabular}{ccc}
\toprule
Layer Name      & Output Size & Size, Channel, Stride                          \\
\midrule
Convolution 1   & $32 \times 32$ & $\left[3 \times 3, 64, 1\right] \times 2$  \\ \hline
Local Pooling1  & $16 \times 16$ & $\left[2 \times 2, 64, 2\right]$          \\ \hline
Convolution 2   & $16 \times 16$ & $\left[3 \times 3, 128, 1\right] \times 2$ \\ \hline
Local Pooling2  & $8 \times 8$   & $\left[2 \times 2, 128, 2\right]$         \\ \hline
Convolution 3   & $8 \times 8$   & $\left[3 \times 3, 256, 1\right] \times 4$ \\ \hline
Local Pooling3  & $4 \times 4$   & $\left[2 \times 2, 256, 2\right]$         \\ \hline
Convolution 4   & $4 \times 4$   & $\left[3 \times 3, 512, 1\right] \times 4$ \\ \hline
Local Pooling4  & $2 \times 2$   & $\left[2 \times 2, 512, 2\right]$         \\ \hline
Convolution 5   & $2 \times 2$   & $\left[3 \times 3, 512, 1\right] \times 4$ \\ \hline
Global Pooling  & $1 \times 1$   & $\left[2 \times 2, 512, 2\right]$         \\ \hline
Fully Connected & $1 \times 1$   & $\left[1 \times 1, 10, 1\right]$           \\
\bottomrule
\end{tabular}
\label{tab1}
\end{table}

\begin{table}
\centering
\caption{ResNet network, modified to suit the CIFAR10 dataset.}
\setlength{\tabcolsep}{3pt}
\begin{tabular}{ccc}
\toprule
Layer Name                 & Output Size    & Size, Channel, Stride                      \\
\midrule
Conv 1              & $32 \times 32$ & $\left[3 \times 3, 64, 1\right]$           \\ \hline
Conv 2-1 + Shortcut & $32 \times 32$ & $\left[3 \times 3, 64, 1\right] \times 2$  \\ \hline
Conv 2-2 + Shortcut & $32 \times 32$ & $\left[3 \times 3, 64, 1\right] \times 2$  \\ \hline
Conv 3-1 + Shortcut & $32 \times 32$ & $\left[3 \times 3, 128, 1\right] \times 2$ \\ \hline
Local Pooling 1            & $16 \times 16$ & $\left[2 \times 2, 128, 2\right]$          \\ \hline
Conv 3-2 + Shortcut & $16 \times 16$ & $\left[3 \times 3, 128, 1\right] \times 2$ \\ \hline
Conv 4-1 + Shortcut & $16 \times 16$ & $\left[3 \times 3, 256, 1\right] \times 2$ \\ \hline
Local Pooling 2            & $8 \times 8$   & $\left[2 \times 2, 256, 2\right]$          \\ \hline
Conv 4-2 + Shortcut & $8 \times 8$   & $\left[3 \times 3, 256, 1\right] \times 2$ \\ \hline
Conv 5-1 + Shortcut & $8 \times 8$   & $\left[3 \times 3, 512, 1\right] \times 2$ \\ \hline
Local Pooling 3            & $4 \times 4$   & $\left[2 \times 2, 512, 2\right]$          \\ \hline
Conv 5-2 + Shortcut & $4 \times 4$   & $\left[3 \times 3, 512, 1\right] \times 2$ \\ \hline
Global Pooling             & $1 \times 1$   & $\left[4 \times 4, 512, 4\right]$          \\ \hline
Fully Connected            & $1 \times 1$   & $\left[1 \times 1, 10, 1\right]$           \\
\bottomrule
\end{tabular}
\label{tab2}
\end{table}

\begin{table*}
\centering
\caption{Classification performances of the evaluated CNNs on the CIFAR10 dataset (averaged over 10 experiments). Best result bold, second-best underlined.}
\setlength{\tabcolsep}{3pt}
\begin{tabular}{cccccc}
\toprule
\multirow{2}{*}{Pooling method}                          & \multicolumn{2}{c}{Pooling Function} & \multirow{2}{*}{Indicator} & \multicolumn{2}{c}{Backbone Network} \\
                                                         & Local pooling    & Global pooling    &                            & VGG               & ResNet           \\
\midrule
\multirow{3}{*}{Standard pooling methods}                & Max              & Max               & V1                         & 93.218            & 94.335           \\
                                                         & Average          & Average           & V2                         & 93.352            & 95.140           \\
                                                         & Stride           & Average           & V3                         & 92.607            & 94.639           \\ \hline
\multirow{3}{*}{Conventional pooling methods \cite{b15}} & Mixed            & Mixed             & V4                         & 93.388            & 94.127           \\
                                                         & Gated-channel    & Gated-channel     & V5                         & 93.197            & 94.548           \\
                                                         & Gated-layer      & Gated-layer       & V6                         & 93.375            & 94.974           \\ \hline
\multirow{5}{*}{Module architecture in proposed methods} & 1 FC             & 1 FC              & P1                         & 93.255            & 95.192           \\
                                                         & 1 FC             & 2 FC              & P2                         & 93.354            & \textbf{95.231}  \\
                                                         & 2 FC             & 2 FC              & P3                         & \underline{93.405}      & \underline{95.201}     \\
                                                         & 1 FC             & Conv              & P4                         & \textbf{93.418}   & 95.199           \\
                                                         & 2 FC             & Conv              & P5                         & 93.260            & 95.163           \\
\bottomrule
\end{tabular}
\label{tab3}
\end{table*}

\subsection{Places2}
\label{5C}


\begin{figure}[htbp]
\centerline{\includegraphics[width=0.9\linewidth]{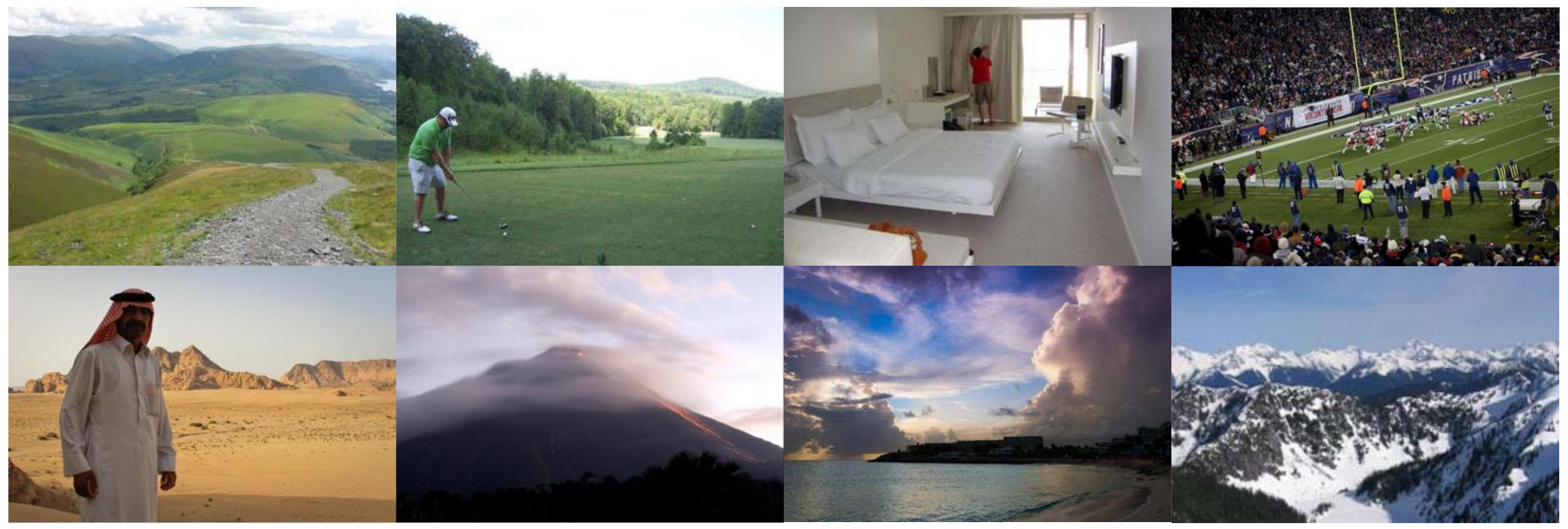}}
\caption{Examples from the Places2 dataset, which contains images of various places.}
\label{fig11}
\end{figure}

The Places2 dataset is a classification dataset containing 365 place classes. The dataset includes 1,803,460 training images and 50 validation images per class. Example images from the Places2 dataset are given in Figure 11. The network was trained by the SGD optimizer and the entire learning period was 120 epochs with a batch size of 256. The initial learning rate, momentum and weight decay of the SGD were set to 0.1, 0.9, and 0.0001, respectively. The learning rate was reduced by one tenth at 30-epoch intervals. The Places2 experiment was conducted in the pre-trained ResNet18 model, and the results are given in Table 2. Clearly, the classification accuracy was improved by the universal pooling.

\begin{table*}
\centering
\caption{Classification performance in the Places2 experiment.}
\setlength{\tabcolsep}{3pt}
\begin{tabular}{ccccc}
\toprule
\multirow{2}{*}{Pooling method} & \multicolumn{2}{c}{Pooling Function} & \multicolumn{2}{c}{ResNet18}      \\
                                & Local pooling    & Global pooling    & top1            & top5            \\
\midrule
Standard pooling methods        & Stride           & Average           & 53.963          & 83.778          \\ \hline
Proposed pooling methods        & 2 FC             & Conv              & \textbf{54.693} & \textbf{84.781} \\
\bottomrule
\end{tabular}
\label{tab4}
\end{table*}

The universal pooling function was qualitatively evaluated by varying the global pooling weights in the universal pooling method. The results of training with the different weights are visualized in Figures \ref{fig12}-\ref{fig15}. As implied in the three cases below, the universal pooling can train various pooling functions, and appropriately selects the pooling function to enhance the classification performance.

\subsubsection{Training by Average Pooling}
\label{5CA}
\begin{figure}[htbp]
\centerline{\includegraphics[width=0.9\linewidth]{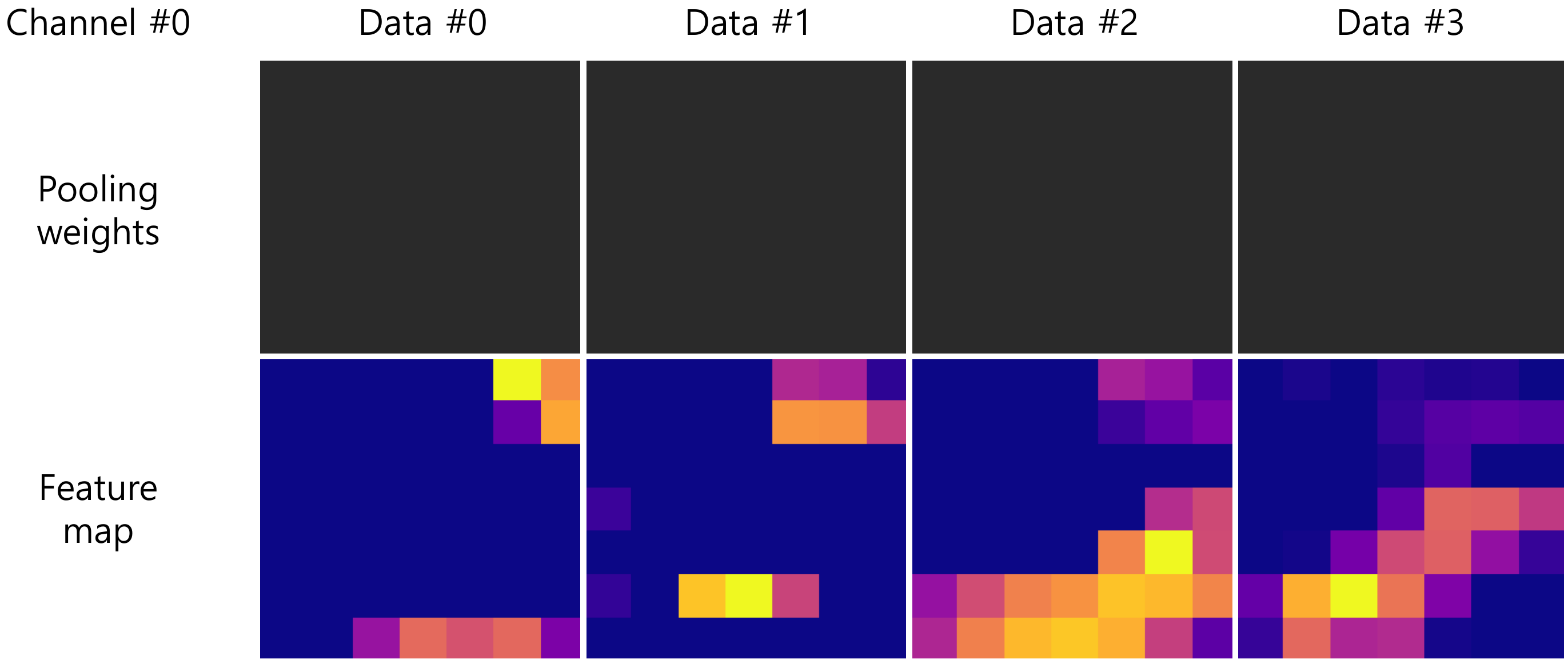}}
\caption{Pooling weights trained by average pooling (top) and the corresponding feature maps (bottom). The trained pooling weights have the same values over the input feature map.}
\label{fig12}
\end{figure}

In almost one-third of the channels (129 out of 512 channels), the pooling weights are the same over the whole feature map, and the pooling reduces to average pooling. The first of these 129 channels is depicted in Figure \ref{fig12}. The trained pooling weights exhibited the same values over the input feature map.

\subsubsection{Training by Flexible Pooling}
\label{5CB}
\begin{figure}[htbp]
\centerline{\includegraphics[width=0.9\linewidth]{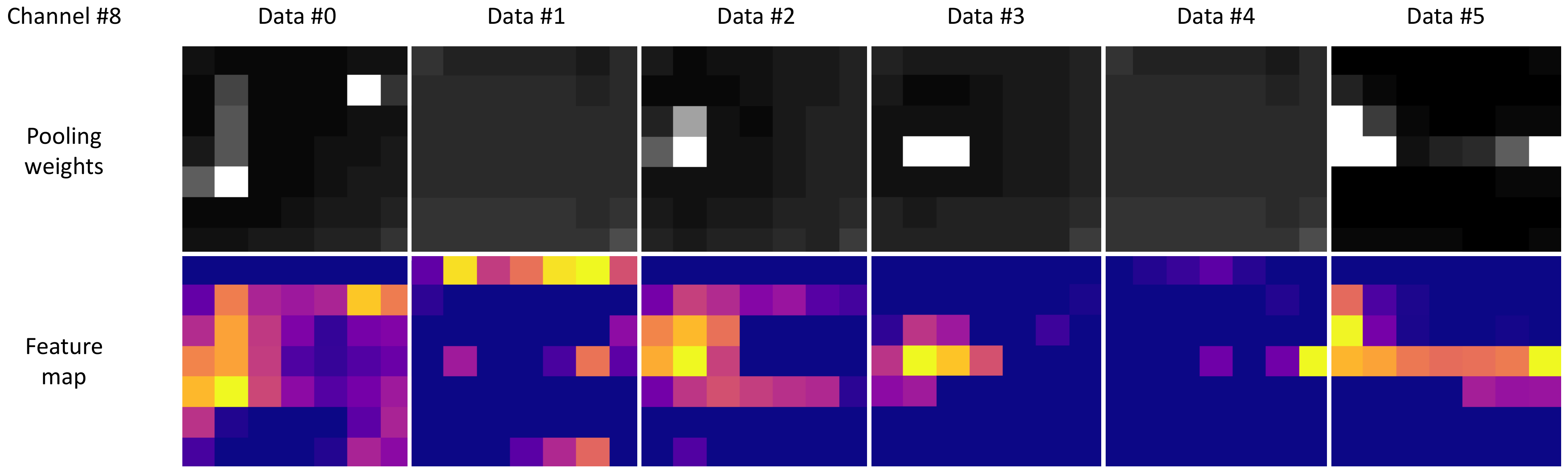}}
\caption{Pooling weights trained by flexible pooling (top) and the corresponding feature maps (bottom). The pooling weights depend on the input feature map. The pooling weights resemble max pooling weights except those of data \#1 and data \#4, which are reminiscent of fixed pooling.}
\label{fig13}
\end{figure}

In almost a fifth of the channels (95 out of 512 channels), the pooling weights depended on the input (see Figure \ref{fig13}). This pooling, called flexible pooling, includes max pooling as a special case. Figure \ref{fig13} shows how the pooling weights change with the feature maps. The trained universal pooling operates similarly, but not identically, to max pooling. In certain data, such as data \#1 and \#4 in the figure, the trained pooling exhibited similar characteristics to fixed pooling (discussed next).

\subsubsection{Training by Fixed Pooling}
\label{5CC}
\begin{figure}[htbp]
\centerline{\includegraphics[width=0.9\linewidth]{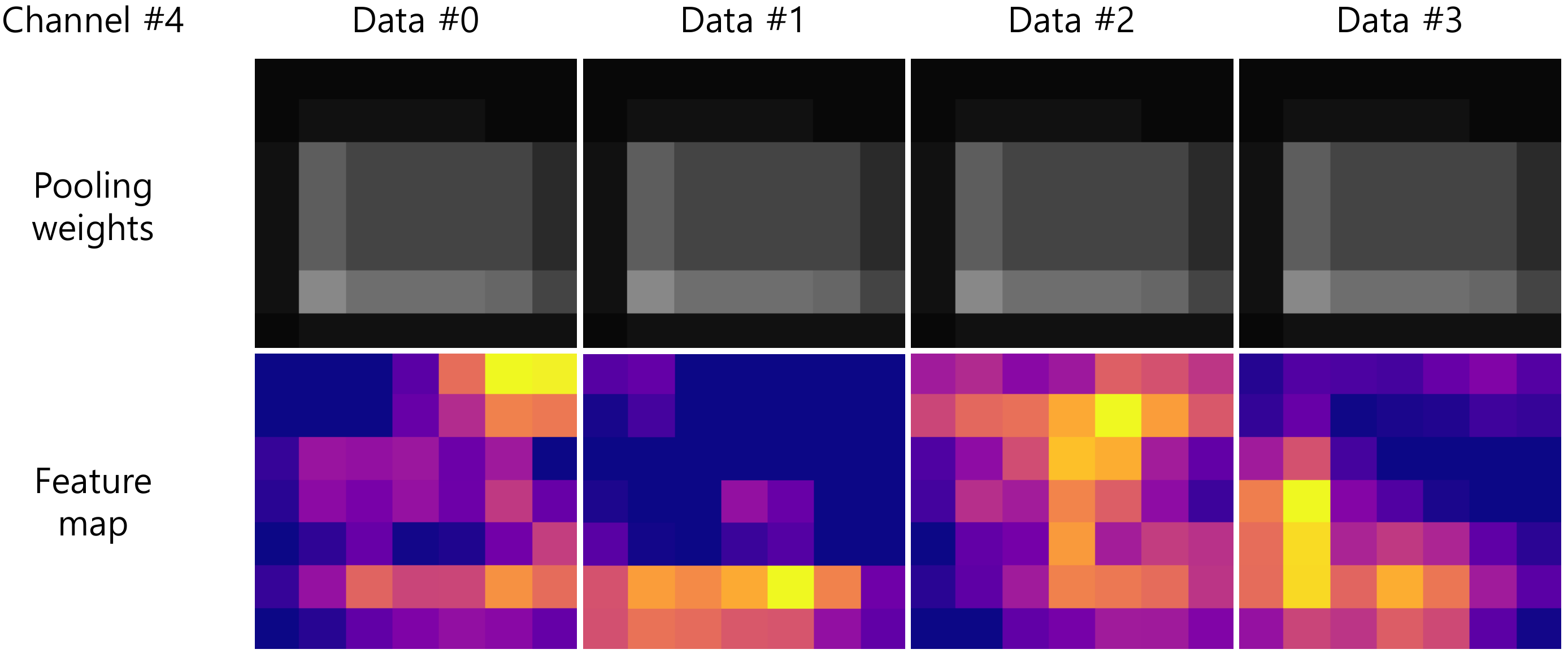}}
\caption{Pooling weights trained by fixed pooling in the proposed method (top) and the corresponding feature maps (bottom). In this channel, the pooling takes the features from the center of the image.}
\label{fig14}
\end{figure}

\begin{figure}[htbp]
\centerline{\includegraphics[width=0.9\linewidth]{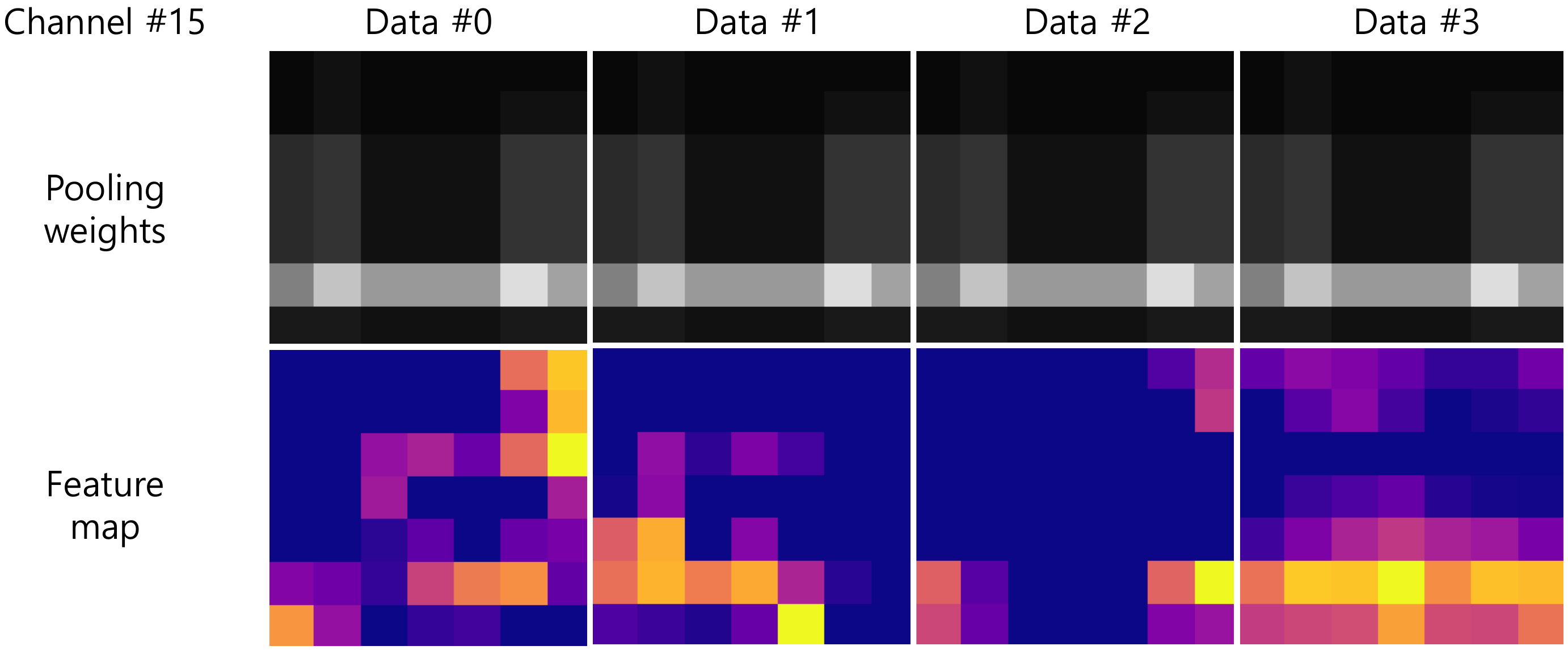}}
\caption{As in Figure \ref{fig14}, but with the pooling features taken from the surrounding area of the image.}
\label{fig15}
\end{figure}

In more than one-half of the 512 channels (288 out of 512 channels), the trained pooling weights were almost fixed, and were nonresponsive to changes in the input feature map. This pooling, called fixed pooling, includes both average and stride pooling. However, average pooling is classified separately from other fixed poolings because the conditions of average and stride pooling differ in the universal pooling. The results of two channels with fixed training weights are visualized in Figures \ref{fig14} and \ref{fig15}. In both cases, the pooling weights depended only on the pooling position in the image, and were independent of the input feature map.

\section{Conclusion}
\label{6}
This paper proposed a new pooling method named universal pooling, which dynamically learns the pooling weights from the given dataset, and which includes the existing popular pooling methods as special cases. Thus, replacing the existing pooling with universal pooling will likely improve the network performance. The proposed pooling was applied to two benchmark datasets and two pretrained models and was validated in comparison experiments.

{\small
\bibliographystyle{ieee_fullname}
\bibliography{My_Paper.bib}

\begin{thebibliography}{10}\itemsep=-1pt

\bibitem{b7}
Long Chen, Hanwang Zhang, Jun Xiao, Liqiang Nie, Jian Shao, Wei Liu, and
  Tat-Seng Chua.
\newblock Sca-cnn: Spatial and channel-wise attention in convolutional networks
  for image captioning.
\newblock In {\em Proceedings of the IEEE conference on computer vision and
  pattern recognition}, pages 5659--5667, 2017.

\bibitem{b10}
Joan~Bruna Estrach, Arthur Szlam, and Yann LeCun.
\newblock Signal recovery from pooling representations.
\newblock In {\em International conference on machine learning}, pages
  307--315, 2014.

\bibitem{b12}
R~Scott Graham, Brian~J Samsell, Allison Proffer, Mark~A Moore, Rafael~A Vega,
  Joel~M Stary, and Bruce Mathern.
\newblock Evaluation of glycerol-preserved bone allografts in cervical spine
  fusion: a prospective, randomized controlled trial.
\newblock {\em Journal of Neurosurgery: Spine}, 22(1):1--10, 2015.

\bibitem{b11}
Caglar Gulcehre, Kyunghyun Cho, Razvan Pascanu, and Yoshua Bengio.
\newblock Learned-norm pooling for deep feedforward and recurrent neural
  networks.
\newblock In {\em Joint European conference on machine learning and knowledge
  discovery in databases}, pages 530--546. Springer, 2014.

\bibitem{b5}
Kaiming He, Xiangyu Zhang, Shaoqing Ren, and Jian Sun.
\newblock Deep residual learning for image recognition.
\newblock In {\em Proceedings of the IEEE conference on computer vision and
  pattern recognition}, pages 770--778, 2016.

\bibitem{b6}
Gao Huang, Zhuang Liu, Laurens Van Der~Maaten, and Kilian~Q Weinberger.
\newblock Densely connected convolutional networks.
\newblock In {\em Proceedings of the IEEE conference on computer vision and
  pattern recognition}, pages 4700--4708, 2017.

\bibitem{b17}
Alex Krizhevsky and Geoffrey Hinton.
\newblock Learning multiple layers of features from tiny images.
\newblock Technical report, Citeseer, 2009.

\bibitem{b2}
Alex Krizhevsky, Ilya Sutskever, and Geoffrey~E Hinton.
\newblock Imagenet classification with deep convolutional neural networks.
\newblock In {\em Advances in neural information processing systems}, pages
  1097--1105, 2012.

\bibitem{b15}
Chen-Yu Lee, Patrick~W Gallagher, and Zhuowen Tu.
\newblock Generalizing pooling functions in convolutional neural networks:
  Mixed, gated, and tree.
\newblock In {\em Artificial intelligence and statistics}, pages 464--472,
  2016.

\bibitem{b13}
Oren Rippel, Jasper Snoek, and Ryan~P Adams.
\newblock Spectral representations for convolutional neural networks.
\newblock In {\em Advances in neural information processing systems}, pages
  2449--2457, 2015.

\bibitem{b1}
Olga Russakovsky, Jia Deng, Hao Su, Jonathan Krause, Sanjeev Satheesh, Sean Ma,
  Zhiheng Huang, Andrej Karpathy, Aditya Khosla, Michael Bernstein, et~al.
\newblock Imagenet large scale visual recognition challenge.
\newblock {\em International journal of computer vision}, 115(3):211--252,
  2015.

\bibitem{b16}
Faraz Saeedan, Nicolas Weber, Michael Goesele, and Stefan Roth.
\newblock Detail-preserving pooling in deep networks.
\newblock In {\em Proceedings of the IEEE conference on computer vision and
  pattern recognition}, pages 9108--9116, 2018.

\bibitem{b4}
Karen Simonyan and Andrew Zisserman.
\newblock Very deep convolutional networks for large-scale image recognition.
\newblock In {\em International conference on learning representations}, 2015.

\bibitem{b3}
Christian Szegedy, Wei Liu, Yangqing Jia, Pierre Sermanet, Scott Reed, Dragomir
  Anguelov, Dumitru Erhan, Vincent Vanhoucke, and Andrew Rabinovich.
\newblock Going deeper with convolutions.
\newblock In {\em Proceedings of the IEEE conference on computer vision and
  pattern recognition}, pages 1--9, 2015.

\bibitem{b8}
Sanghyun Woo, Jongchan Park, Joon-Young Lee, and In So~Kweon.
\newblock Cbam: Convolutional block attention module.
\newblock In {\em Proceedings of the European Conference on Computer Vision
  (ECCV)}, pages 3--19, 2018.

\bibitem{b19}
Matthew~D Zeiler and Rob Fergus.
\newblock Stochastic pooling for regularization of deep convolutional neural
  networks.
\newblock In {\em International conference on learning representations}, 2013.

\bibitem{b9}
Shuangfei Zhai, Hui Wu, Abhishek Kumar, Yu Cheng, Yongxi Lu, Zhongfei Zhang,
  and Rogerio Feris.
\newblock S3pool: Pooling with stochastic spatial sampling.
\newblock In {\em Proceedings of the IEEE conference on computer vision and
  pattern recognition}, pages 4970--4978, 2017.

\bibitem{b14}
Duo Zhang, Erlend~Skullestad Holland, Geir Lindholm, and Harsha Ratnaweera.
\newblock Enhancing operation of a sewage pumping station for inter catchment
  wastewater transfer by using deep learning and hydraulic model.
\newblock {\em arXiv preprint arXiv:1811.06367}, 2018.

\bibitem{b18}
Bolei Zhou, Agata Lapedriza, Aditya Khosla, Aude Oliva, and Antonio Torralba.
\newblock Places: A 10 million image database for scene recognition.
\newblock {\em IEEE transactions on pattern analysis and machine intelligence},
  40(6):1452--1464, 2018.

\end{thebibliography}
}

\end{document}